# An Iterative Convolutional Neural Network Algorithm Improves Electron Microscopy Image Segmentation

### Abstract

To build the connectomics map of the brain, we developed a new algorithm that can automatically refine the Membrane Detection Probability Maps (MDPM) generated to perform automatic segmentation of electron microscopy (EM) images. To achieve this, we executed supervised training of a convolutional neural network to recover the removed center pixel label of patches sampled from a MDPM. MDPM can be generated from other machine learning based algorithms recognizing whether a pixel in an image corresponds to the cell membrane. By iteratively applying this network over MDPM for multiple rounds, we were able to significantly improve membrane segmentation results.

## 1    Introduction

To further expand our understanding on the structure and working mechanisms of the human brain, it has been deemed necessary to map the entire neural connections of the nervous system at the micro-scale level. One main approach is to acquire serial 2-D images of brain tissue at nanometric resolution with serial-section Transmitted Electron Microscopy (ssTEM). To generate a connection map in 3-D, an expert is then needed to manually analyze those 2-D images to segment different neural tissues in the image [6;7;21]. This task is only realistic if the volume of images that needs to be processed is reasonably small. However, that usually is not the case. For example, a typical image set of the V1 area of a mouse brain with resolution of $4\times4\times45$ nm produces 12TB of data[5]. Processing this amount of data manually becomes quite impractical.

Much effort has been put to develop tools to automatically process those images. Previous approaches [8;14;16] use the contextual information surrounding a pixel to give a probability value to distinguish the cell membrane. Applying those detectors to every pixel of an original EM image leads to a pixel wise MDPM.

Post-processing on top of these detection maps is necessary in order to obtain the final region segmentation results. The post-processing can be simple, for example, simple thresholding, or creating a smooth probability map by using a median filter [8]. Other more graphical models based on region merge/clustering techniques such as superpixel [10] and watershed merge [18;19]have also been applied on this or similar tasks to refine segmentation on top of a detection map.

In this work, we introduce a new algorithm without resorting to any kind of graphical model based region merge/clustering algorithms. Simply by training a Convolutional Neural Network (I-CNN) to recover the masked out center pixel value of patches sampled from MDPM, and then iteratively applying this network over MDPM, one can obtain a high quality segmentation map both visually and measured by the foreground-restrict Rand score [1;13]. The algorithm significantly improves the foreground-restrict Rand score over the original MDPM map and achieves a similar level of performance to the current leading algorithms on ISBI 2012 EM segmentation challenge dataset.

## 2    Related works

Computer vision research on real world image contour detection and segmentation tasks has come up with many solutions to ensure the consistency of image segmentation. A good number of them use some kind of over-segmentation to obtain super-pixel like pre-segmentation [3;10]. This process is then followed by graphical models to merge sub-regions and reach a final segmentation map. Models such as tree merge, Conditional Random Fields (CRFs), Markov Random Fields (MRFs) or graph cuts algorithm have been used for this purpose [10;12;17;22].

For the EM image segmentation task, the detection of membrane resembles the contour detection problem in general computer vision. The quality of membrane detection can be directly measured by the pixels error, but just like in contour detection, a high quality membrane detection does not guarantee a good segmentation [4]. A small gap in the contour formed by the detected membrane can lead to an incorrect merge of two different regions. Or a false section of membrane can incorrectly split one region into two parts. This is especially important for neural circuit reconstruction, as one need to identify every neurite to re-link them in 3-D. The quality of the segmentation can be measured by the Rand index[13], which measures the resemblance of two clusters of segmentations. Specifically, the ISBI 2012 EM segmentation challenge dataset uses the fore-ground restrict Rand error index as one leading index to score different algorithms [1].

In a typical MDPM such as the one from Ciresan's work [8] the detection probability of every pixel is made in isolation, which increases the possibility of having gaps in the boundaries. In Ciresan's work they used a simple median filter of small radius to smooth out the detection map. This procedure gained them a state-of-the-art level of result back in 2012. However, the limitation of applying a median filter is its isotropy; that is, if one applies a median filter of increasing radius to the MDPM the performance will quickly deteriorate as the radius becomes larger. Therefore, an algorithm that can not only take long range information into consideration to smooth out the map but that it is also able to avoid isotropy smoothing was needed. To avoid this limitation, we have developed an Iterative Convolutional Neural Network (ICNN) that significantly improves the definition of boundaries.

Like in natural image segmentation, several authors have proposed a graphical model algorithm to merge regions starting from MDPM. For example Andres and later Liu [2;19] trained a classifier which merges over-segmented regions generated from water-shed over segmentation. Funke *et al*. [11] used a tree structure for segmentation hypothesis and reach segmentation with MAP inference in a CRF. Unlike their work, we did not train for the target to reduce segmentation error. What our network does is to simply enforce constraints to push MDPM toward the manifold of the membrane structure distribution that it learned through recovering missing information. While this new algorithm did not rank at first place on the Rand error score, it performs significantly better than all top leading groups on the other two scores.

Our approach is also related to the image super resolution convolutional neural network [9] where a convolutional neural network is trained to restore high resolution from low resolution images, and to de-noise network [15] where the purpose is to restore a noisy image. If we only think about a single iteration, we use a similar convolutional neural network architecture to those two networks, where their training targets can all be thought of as to recover original pixel values from distorted images either caused by noise or blurring. Our network is different in that it applies multiple iterations instead of only modifying a single pixel in isolation. Multiple iterations give our method the power to turn on/off a whole section of membrane (see Figure 2) resulting in a great improvement in boundary consistency.

## 3    System description and result

### 3.1    Dataset description

The dataset used in this experiment consists of two stacks of EM images used in the ISBI 2012 EM segmentation challenge. One stack has 30 EM images and their corresponding labels for training. The other stack also contain 30 images, their labels are concealed.

## 3.2 Base convolutional network training

The network used in this part is analogous to the convolutional network implemented by Ciresan [8]. However, it can be replaced by any other algorithm that generates a high quality membrane detection probability map. To differentiate the two networks, we call this base network CNN and the Iteration Convolutional Neural Network I-CNN from here on. As in the work of Ciresan [8], CNN is trained to yield the probability of whether the pixel at the center of a given image patch (65x65 size) is a membrane pixel. The difference is that our network has 4 convolutional layers, one fully connected layer and one softmax layer. Rectified linear activation functions are used in the network except in the last layer, where a softmax function is used. Also, similar to the work of Ciresan[8], we augment the dataset by rotation, flipping and mirroring images in a total of 8 different transformations. One example is shown in the original probability map image of Figure 1. A detailed network architecture and hyper parameter settings will be made available in the code that will be released with the final version of the paper.

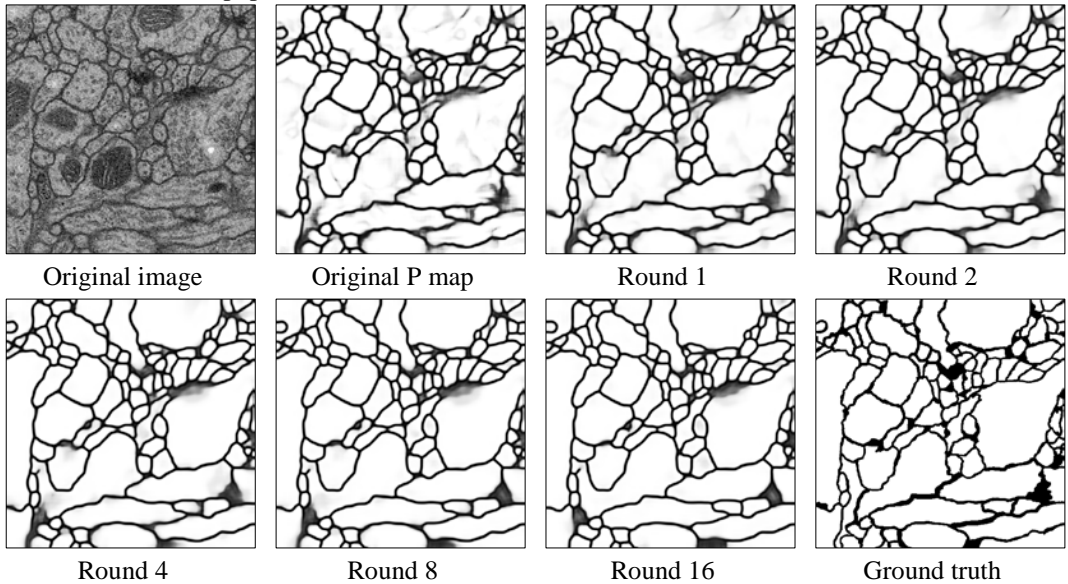

Original image    Original P map    Round 1    Round 2

Round 4    Round 8    Round 16    Ground truth

Figure 1. One sample image processed by I-CNN. Probability map is shown in reverse manner. The darkest pixel value means that the probability of being membrane is 1.  Original image is a raw EM image; Original P map is the MDPM output from the base CNN network. Round 1, 2, 4, 8, 16 are the maps processed by I-CNN 1, 2, 4, 8, 16 rounds, respectively. Ground truth is the map labeled by a human expert, black indicates a membrane pixel, and white indicates a non-membrane pixel.

To test the performance of this network, a network was trained with patches extracted from 25 images from a training set where 5 images were left out as a validation set. As reported by Ciresan [8], we sampled the example patches evenly from both membrane and non-membrane classes for the training of both CNN and ICNN networks. Also similar to Ciresan work [8], we trained a bias correction curve to correct for bias caused by this sampling process. This correction curve is used throughout the whole work to correct for sampling bias. After the validation set likelihood stops to improve, we save the network parameters where we obtain the best likelihood loss on the validation set for the next step. The trained network is then applied on the images from the test set to generate the probability map of those images. For better precision, the test images are also processed in 8 different transformed images and resulting maps are averaged to obtain one single probability map. The resulting map is submitted to the ISBI challenge 2012 server for evaluation; the score is shown in Table 1 compared against the congregated result (averaged results from multiple networks output) published by Ciresan [8]. Although our score is slightly lower, our performance is better than each of their single networks. The probability map generated from this network is used for the next stage. Since this part is not the focus of this work, we did not attempt to implement multiple networks of different architectures for better precision [8].

Table 1 Network performance comparison

| Network | Rand error | Warping Error | Pixel Error |
|---|---|---|---|
| Convolutional network in [8] | 0.0483 | 0.000434 | 0.0604 |
| Our base network | 0.0602 | 0.000426 | 0.0627 |
| Boosted base network | 0.0551 | 0.000431 | 0.0646 |
| **Our I-CNN (currently #4)** | **0.0263** | **0.000509** | **0.0747** |
| #1 Leading group | 0.0178 | 0.000711 | 0.154 |
| #2 Leading group | 0.0189 | 0.000669 | 0.155 |
| #3 Leading group | 0.0228 | 0.000808 | 0.110 |
| #5 Leading group | 0.0281 | 0.00107 | 0.155 |
| The scores list on the ISBI challenge public leaders board are the highest score that every group achieved across all their submission per index. Here we list the score from their best ranked submission. | | | |

### 3.3    Generation of MDPM for training set images

In the next section, we will describe our approach to refine the probability map for better segmentation accuracy. The idea here is to train another convolutional network, which can refine the raw MDPM in order to obtain better segmentation performance. For this task, we use the raw MDPM with its corresponding ground-truth segment map to train the network. In order to train an I-CNN that can refine a MDPM to reach the final segmentation map, we need the MDPM we can train with. If we apply the CNN described in the last section to the training image set that was used to train the CNN, maps will have severe overfitting, and thus, cannot be used for training. To overcome this, we performed a ten-fold split on the training set so the network is trained with ten different versions of splits each containing 27 training images and 3 left out images. After training each model, the procedure is applied on their corresponding 3 left out images to generate the corresponding probability maps. In this manner, we collect 30 MDPMs that do not have an overfitting issue. As a by-product, each model is also applied to the 30 testing images to obtain their MDPM stack, and ten versions of probability MDPM stack are averaged to obtain one MDPM stack of slightly better quality (score shown in table 1). This averaged test set MDPM stack is used in the next step.

### 3.4    Iterative convolution neural network

The pixel wise likelihood map generated from the network described in the last section showed high pixel wise accuracy yet it was short of local consistency in certain areas (see Figures 1 & 2), which is inconsistent with the spatial continuity of the cell membrane. A human labeler with the implicit model of the cell membrane spatial continuity in mind will integrate contextual information and conform to the spatial continuity constraint. In the approach described in the last two sections, although contextual information is used for generating pixel detection probability values, those probability values are generated independently. This may leave gaps in a continuous boundary and lead to an eventually low quality connectome. On the other hand, we want to let longer range information outside the patch window (in our typical case 65x65 pixels) to propagate into the window and improve the detection probability.

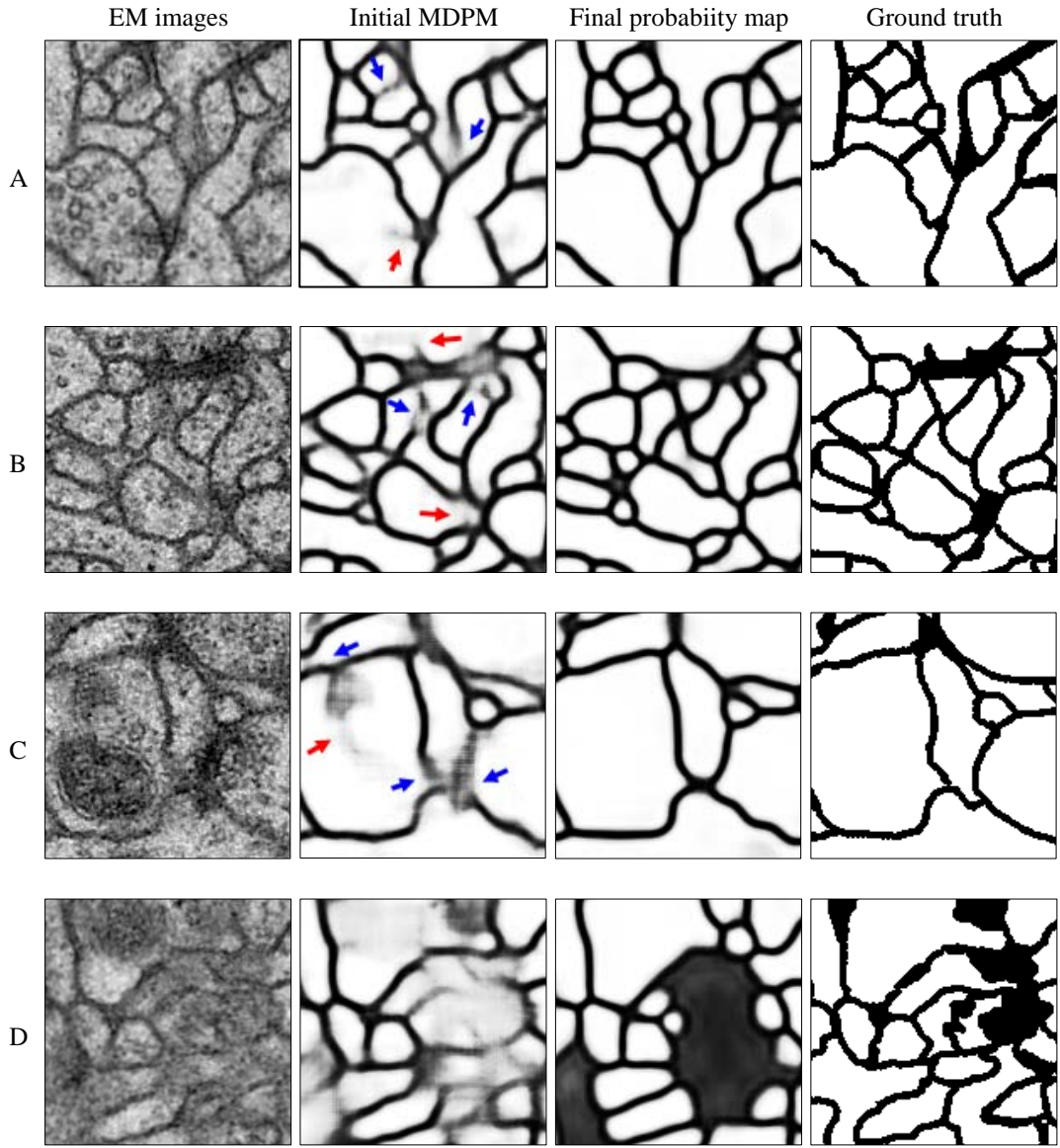

Figure 2. Examples of patches where the model closes gaps and removes uncertain membrane sections. Blue arrows indicate where the model adds or solidifies a section of link between membrane parts; red arrows indicate where the model removes sections of uncertain membrane. As we can see the model does not always make the right decision. In D, the model failed badly because it was not able to properly process the distorted map that was possibly caused by an uneven cut during tissue slice preparation.

Here we propose a simple convolutional network (I-CNN) which directly learns the statics of segmentation maps to significantly improve the segmentation quality. The main difference between this network and the previous network is the input images. In the earlier CNN network, the input to the network is the raw EM image. In this case, the input image patches are replaced by the patches extracted from MDPM, while the label of the center pixel is removed. The training target stays the same as the label of the center pixel from the ground truth map. Since the structure used is quite similar as to the CNN network, it will not be described here. All detail and code will be made available with the final version of the paper. The same training procedure as in 3.2 is used for this network. We also resample the patch to maintain balanced class as in 3.2 and the correction curve in 3.2 is again used here to correct sampling bias. This step will lead to slightly reduced pixel error as revealed by a smaller likelihood loss value at the end of the training on the validation dataset (0.228 vs. 0.239). As we stated earlier, we try to improve segmentation by integrating longer range information that is included in the neighboring pixel detection probability. On the other hand, when we apply this network to a MDPM we tend to push every pixel in the new map toward a higher likelihood given its neighborhood. This leads to another important change; the detection probability of neighboring pixels is not generated in isolation anymore. What will happen if you repeatedly apply the same I-CNN on the membrane detection probability refined by the last round of the same I-CNN?

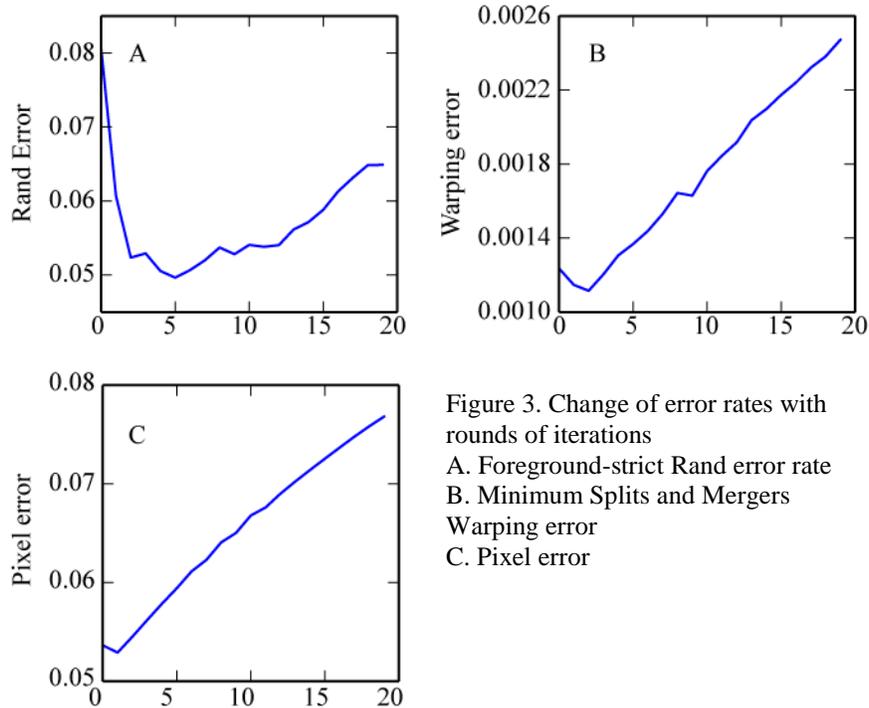

Figure 3. Change of error rates with rounds of iterations
A. Foreground-strict Rand error rate
B. Minimum Splits and Mergers Warping error
C. Pixel error

## 3.5   Result

As shown in Figure 1, when we iteratively apply I-CNN to the MDPM, the first thing we notice is that by iteratively refining the probability, we gradually remove the noise in the map. After 16 rounds, the map turns out to be a map with rather clear boundaries vs. fuzzy boundaries in the original MDPM. More importantly, if we zoom into areas where CNN cannot make an affirmative inference about pixels, as shown in Figure 2, we can see that I-CNN is able to integrate information in a neighborhood to recognize (blue arrows) a section of membrane shown with low probability but with good spatial continuity, and eventually labels the section with high confidence and closes gaps at the boundary. At the same time I-CNN can identify noise pixel and areas (Red arrow) that do not appear like a section of membrane and eventually completely remove it. Yet, it can make mistakes in areas of great uncertainty, for example, in the case shown in Figure 2D.

We then measured the segmentation result by the standard score used by ISBI EM segmentation challenge. For this part of experiment, two different I-CNNs are trained each with 25 MDPM and grounding truth label map, and 5 MDPM are left out as validation set from training. Different 5 subsets are used for those two networks. After training, I-CNN is applied to their corresponding left out MDPM set for a total of 20 rounds; their segmentation error scores are then measured for every round by the code provided by the challenge website[1]. Resulting scores are then averaged across ten maps and shown in Figure 3. Surprisingly, even though originally we train our network only to recover pixel error, we can observe that by iteratively applying I-CNN on the membrane detection map one can dramatically reduce the Rand error of the segmentation result at the beginning of the iteration. This reduction in Rand error disappears only after about 6 rounds and afterwards deteriorates. At the same time, as expected it indeed reduces the pixel error (Figure 3-C) and warping error (Figure 3-B) rates at the very beginning of the iteration. Although the effect reversed in the $2^{nd}$ round, this is not very important for the neural circuit reconstruction task. We also applied the I-CNN to the test image stack for 6 rounds and submitted the result to ISBI 2012 EM segmentation challenge website obtaining a Rand error score of 0.0263, which is much better than the score of 0.0551 obtained from the original CNN result before refining (Full score comparison in Table 1).

## 4  Conclusion and Discussion

The new algorithm presented in this work learns the manifold of membrane morphology distribution; it enforces these constraints through iteration on a MDPM, refining it to fit a membrane morphology distribution learned from the training data. Through this process, a much better neurite segmentation results can be obtained. In other perspective, instead of generating a membrane detection probability of every pixel in isolation, through applying I-CNN iteratively over MDPM, we congregated information in the local neighborhood and obtained significantly better consistency in neighboring pixels. A great improvement, measured by Rand error, was achieved over the original MDPM result. The resulting score is competitive with the state-of-the-art result even when we did not start with the MDPM result with much better precision as the authors of the current state-of-the art holder did. Furthermore, we achieved much better scores on two other metrics (see Table 1).

Currently, we apply networks to images pixels wisely at the iteration stage, which causes a large amount of redundant computation on neighboring pixels. Long *et al.* [20] and other groups, developed full convolutional neural network that can avoid this redundant computation cost by converting a patch-wise trained convolutional neural network to a network that is able to efficiently process a whole image. In the future, we can apply this technique to our approach for more efficiency at the iteration stage. A small modification to our training stage, where instead of replacing the center pixel label with a fixed value of 0, we inject random noise to the center pixel so that the network will learn to ignore this pixel. In this way, we can take advantage of the efficiency of full convolutional neural network.

Bearing in mind that the continuity of the cell membrane extends beyond a 2-D plane, it is shown that adding 3-D information from neighboring planes will likely improve the segmentation accuracy [11;18], especially in the regions of low confidence. Expanding this algorithm to include 3-D information is straight forward; however, it will require more data for training. Given the enormous volume of data that are involved in generating the connectome, it does make sense to enable this approach to gain more accuracy in the future.

In the EM segmentation, what our I-CNN network learns is relatively simple statistical structures. It would be even more interesting if we can apply the same architecture to other tasks, for example, to generate realistic looking natural images or to regenerate an image from a partially damaged photo to have a natural looking.